\documentclass{article}
\usepackage{ijcai22}

\usepackage{times}
\usepackage{soul}
\usepackage{url}
\usepackage[hidelinks]{hyperref}
\usepackage[utf8]{inputenc}
\usepackage[small]{caption}
\usepackage{graphicx}
\usepackage{amsmath}
\usepackage{amsthm}
\usepackage{booktabs}
\usepackage{algorithm}
\usepackage{algorithmic}
\title{Conditional super-network weights}

\author{
	Kevin A. Laube
	\and
	Andreas Zell
	\affiliations
	University of Tübingen, Germany
	\emails
	\{kevin.laube, andreas.zell\}@uni-tuebingen.de
}

\usepackage{graphicx}
\usepackage{amsmath}
\usepackage{amssymb}
\usepackage{caption}
\usepackage{subcaption}
\usepackage{booktabs} 
\usepackage{multirow}
\usepackage{enumitem}

\def\dcifar100{CIFAR100}
\def\dcifar10{CIFAR10}

\def\bnb201{NAS-Bench-201}

\def\mtab{Lookup Table}

\begin{document}

\maketitle

\begin{abstract}
	Modern Neural Architecture Search methods have repeatedly broken state-of-the-art results for several disciplines. The super-network, a central component of many such methods, enables quick estimates of accuracy or loss statistics for any architecture in the search space.
	They incorporate the network weights of all candidate architectures and can thus approximate specific ones by applying the respective operations.
	However, this design ignores potential dependencies between consecutive operations.
	We extend super-networks with conditional weights that depend on combinations of choices and analyze their effect.
	Experiments in NAS-Bench 201 and NAS-Bench-Macro-based search spaces show improvements in the architecture selection and that the resource overhead is nearly negligible for sequential network designs.
\end{abstract}

\section{Introduction}  
\label{p4_intro}

Owed to the promises of improving over hand-designed networks and reducing manual effort, Neural Architecture Search (NAS) has been drawing attention in academia and industry alike.
Where early works employed reinforcement learning (RL) \cite{zoph2016neural,zoph2018learning} or evolutionary algorithms (EA) \cite{real2018regularized} to guide the training of thousands of models on hundreds of GPUs over days,
the invention of the one-shot approach \cite{pham2018efficient} enabled NAS in a reasonable time, even on a single GPU.

A variety of further one-shot methods followed, not only limited to RL \cite{proxylessnas}, EA \cite{guo2020single} or bayesian optimization \cite{shi2019bridging,white2019bananas}, but also including especially popular gradient based approaches \cite{liu2018darts,xie2018snas,dong2019gdas,proxylessnas,stamoulis2019singlepath}.
As different as the methods may be, all of them require that the trained one-shot model enables drawing valid conclusions about the search space it was built from.
The study of this consistency is a growing research trend \cite{EvalSeachPhase2019,yu2020train,FairNAS,li2020blockwisely,peng2020cream}, promising to improve the quality of NAS results in every search space.

We present \textit{conditional super-network weights}, a modification to the one-shot model design that enables candidate operations in consecutive layers to specialize towards each other.
Due to a process of \textit{weight splitting}, these weights can be trained efficiently and with little overhead.
Experiments in four search spaces of NAS-Bench 201 and NAS-Bench-Macro show that the modified models become capable of selecting considerably better architectures than their baselines.

\begin{figure}[t]
	\begin{center}
		\includegraphics[width=1\linewidth, trim=170 1250 310 155, clip]{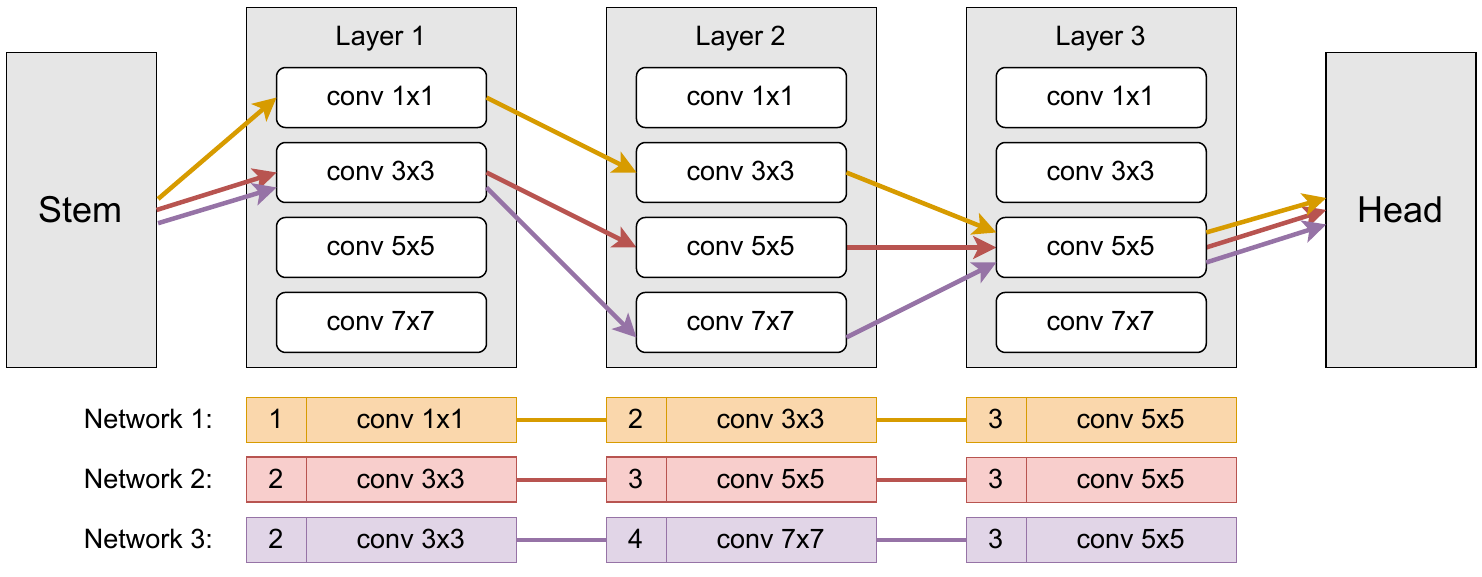}
	\end{center}
	\vskip-0.3cm
	\caption{
		A small sequential super-network with three layers and four candidate operations in each.
		The connecting arrows constitute three specific architectures in the search space, subsets of the over-complete computational graph.
		All of them use the \textit{conv 5}$\times$\textit{5} operation in the last layer and thus share its weights. Aside from Layer~2, Networks~2 and~3 are identical.
	}
	\label{fig_p4_super}
\end{figure}

\section{Foundations and Related work}
\label{p4_rel}

\paragraph{Neural Architecture Search (NAS)}
Following the initial success of \cite{zoph2016neural}, the automated design and discovery of network architectures has become a topic of substantial academic and industrial interest.
The challenge of NAS is to find the optimal architectures in the search space $\mathcal{A}$, which may contain many billion candidates.

\paragraph{Super-networks}
Early works evaluated the considered architectures after training them from scratch, spending thousands of GPU hours \cite{zoph2018learning,real2018regularized}.
In contrast, many modern approaches train a single super-network (also called \textit{one-shot} or \textit{over-complete}) to efficiently predict the performance of any architecture in the search space \cite{pham2018efficient}.
This super-network model is over-complete since it contains all considered candidate operations and, therefore, all candidate networks.
Any specific architecture can be trained or evaluated by selecting the respective operations.
An example is visualized in Figure~\ref{fig_p4_super}.
The colored arrows illustrate three different candidate architectures that share some weights using the same operations.

In this example, any candidate network is acceptable if it contains precisely one operation per layer.
Since the number and order of layers and their candidate operations are clearly defined here, Networks~2 and~3 can be uniquely identified with the descriptions $Net_{(2, 3, 3)}$ and $Net_{(2, 4, 3)}$.

\paragraph{Single-Path One-Shot (SPOS)}
Unlike many other NAS methods, SPOS \cite{guo2020single} decouples the architecture search from the super-network training.
For every batch, candidate operations are picked uniformly at random. For example, three consecutive batches could use the candidate networks $Net_{(1, 2, 2)}$, $Net_{(2, 4, 1)}$, and finally $Net_{(2, 3, 4)}$.
Thus trained, the super-network is viewed as an unbiased collection of individually trained architectures.
Any specific architecture can then be evaluated on a validation dataset by choosing the corresponding operations without further training.

These estimates guide proven hyper-parameter optimization techniques for discrete values in the subsequent architecture search.
\citeauthor{guo2020single} use a simple evolutionary algorithm to maximize the network accuracy under a FLOPs constraint.

\paragraph{Quantifying NAS}
A comprehensive evaluation of a NAS algorithm requires knowing the ground-truth accuracy or loss values of every architecture in the given search space $\mathcal{A}$.
Such information is generally only available in NAS benchmarks (\cite{ying19nasbench101,transnas101,siems2020nasbench301} and more),
where small spaces of a few thousand architectures are evaluated for precisely this purpose.
Our evaluation is based on two such benchmarks, NAS-Bench-201 \cite{dong2020nasbench201} and NAS-Bench-Macro \cite{nbmacro}.

\section{Method}
\label{p4_method}

\subsection{The problem}
\label{p4_method_problem}

The super-network serves as a cheap evaluation model substituting all stand-alone networks in its search space to reduce their immense combined training costs.
As seen in Figure~\ref{fig_p4_super}, its weights are shared by the different candidate architectures.
In particular, the three example networks use the 5$\times$5 Convolution in Layer~2.
However, does it make sense to use the same weights for this operation, no matter what comes before or after?

Formally, denote $O_{(2, 3)}$ as the third candidate operation (5$\times$5 Convolution) in Layer~2.
The example Network~3, $Net_{(2, 4, 3)}$, is thus uniquely defined by the set of its candidates $\{O_{(1, 2)}, O_{(2, 4)}, O_{(3, 3)}\}$.
Any $O_{(x, y)}$ uses the same weights no matter which particular candidate architecture is currently used.
$O_{(3, 3)}$ is thus part of all three example networks in Figure~\ref{fig_p4_super}.
A logical consequence is that all candidate operations in the second layer $\{O_{(2, 1)},..., O_{(2, n)}\}$ need to produce structurally similar information, otherwise $O_{(3, 3)}$ could not function properly.
However, the candidates $\{O_{(2, 1)},..., O_{(2, n)}\}$ may have different complexity and capacity. In this example, they differ only by their convolution kernel sizes. Still, for a 1$\times$1 and a 7$\times$7 convolution to produce similar outputs, the latter must not use most of its capacity.
Furthermore, every operation in the third layer $\{O_{(3, 1)},..., O_{(3, n)}\}$ must adapt to the similar outputs of any $\{O_{(2, 1)},..., O_{(2, n)}\}$.
To summarize the problem: All candidates of any one layer must adapt to similar inputs and produce similar outputs. In the worst case, the candidates with the lowest capacity limit the intermediate network information.

In practice, several works find a substantial performance disparity between architectures that are trained independently and their equivalent subsets in a super-network (e.g. \cite{chu2020fair,peng2020cream,li2020blockwisely,zhao2020fewshot}).
Nonetheless, super-networks are a central component of many state-of-the-art methods such as \cite{cai2019onceforall,peng2020cream,attentivenas,hardcore}.

\subsection{Conditional super-network weights}
\label{p4_method_add}

As just described, the weight sharing of the super-network paradigm encourages questionable co-adaptions of all candidate operations in the same layers.
\cite{sharinginsights} find that decreasing the degree of weight sharing improves the search results but substantially increases the costs.
The presented \textit{conditional super-network weights} approach the problem from a different perspective by decreasing the pressure for candidate operations to co-adapt.
The fundamental idea of the approach is to give each candidate operation the ability to behave differently depending on which operation generated its input.
$O_{(3, 3)}$ in Figure~\ref{fig_p4_super} should be able to behave differently depending on which $O_{(2, x)}$ is part of the current network.
Such pair-wise specializations are desired between any candidates in subsequent layers $(O_{(n, x)}, O_{(n+1, y)})$.

\begin{figure}
	\centering
	\includegraphics[height=6cm, trim=280 440 20 110, clip]{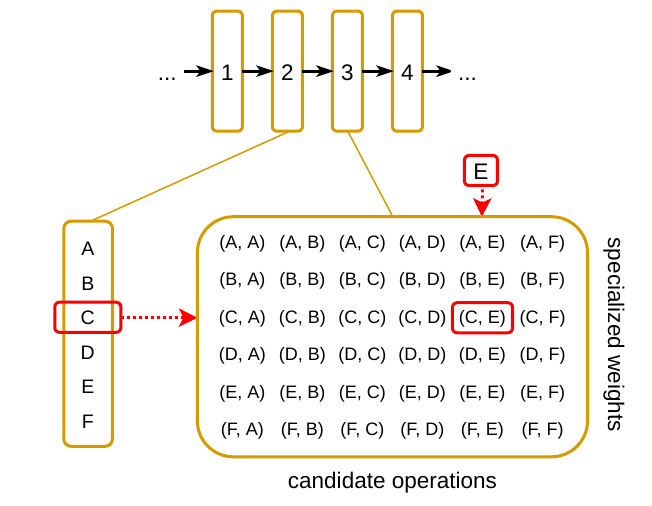}
	\vskip -0.3cm
	\caption{
		An example of conditional weights in a~purely sequential super-network that has six candidate operations (A to F) in each layer. Every candidate has a different set of weights for each candidate in the previous layer. In this particular forward pass, layer~2 is set to operation~C, and layer~3 to~E (marked red).
		The used operation in layer~3 is thus~$O_{(3, (C, E))}$.
	}
	\label{fig_p4_aw}
\end{figure}

A simple approach to achieve this specialization is by choosing the weights of every candidate operation in a topology-aware way.
More precisely: every single candidate has different sets of weights, which are tied to the candidate operations in the layer before.
$O_{(3, y)}$ does not depend on the previous layer, but $O_{(3, (x, y))}$ does.
While the previous layer does not affect the type of operation (e.g., a 5$\times$5 Convolution), its specific weights are finetuned individually.
This is visualized in Figure~\ref{fig_p4_aw}.

Due to the complexity of this approach, there are two significant concerns:
Firstly, the number of super-network weights has effectively been multiplied by the number of candidate operations. Consequentially, given the same amount of training as a regular super-network, the weights will not adapt well.
An increased training time would address this issue but is naturally undesired.
Secondly, while the specialized weights of each candidate (e.g., $\{O_{(3, (A, C))}, ..., O_{(3, (F, C))}\}$) are not supposed to be identical, it is unlikely that they should be vastly different. However, if they are trained entirely independently, this is likely to be the case.
Both concerns can be addressed with a simple approach that we call \textit{weight splitting}:
Until a specific training epoch, say at three-quarters of the allocated time, the specialized weights are disabled and the super-network trained as usual.
Only then are the weights of all operations $O_{(x, y)}$ \textit{split}, i.e., copied once for each candidate operation in the previous layer. A set $\{O_{(x, (A, y))}, ..., O_{(x, (F, y))}\}$ is created from every candidate $O_{(x, y)}$, to enable specialization towards the prior operations A~to~F for the remainder of the training.
Since all weight sets are initially trained as one, they are similar to one another and require no additional training time. As a disadvantage, the choice of when to split each candidate is an open question.

While Figure~\ref{fig_p4_aw} only presents the case of a fully sequential network, the generalization to multiple paths is trivial:
If candidate $O_{(n+1, y)}$ depends on multiple previous operations $\{O_{(n, x_1)}, ..., O_{(n, x_m)}\}$, then the set of weights that considers all of them fairly is $O_{(n+1, (x_1, ..., x_m, y))}$.
Although the increased number of weights may appear alarming in memory consumption, especially for multi-path networks, the actual impact during training is not that significant.
The reason is that the network weights themselves require much less memory during training than storing the intermediate tensor results for backpropagation.
Detailed resource statistics are evaluated in Section~\ref{p4_exp_resources}.

\subsection{Search spaces}
\label{p4_method_datasets}

We evaluate super-networks modified with conditional weights in four different search spaces from the following two NAS benchmarks:

\subsubsection*{NAS-Bench-201}

\begin{figure}[t]
	\begin{minipage}{0.45\linewidth}
		Candidate operations:
		\begin{itemize}[noitemsep,parsep=0pt,partopsep=0pt]
			\item Zero
			\item Skip
			\item 1$\times$1 Convolution
			\item 3$\times$3 Convolution
			\item 3$\times$3 Average Pooling
		\end{itemize}
	\end{minipage}
	\hfill
	\begin{minipage}{0.52\linewidth}
		\includegraphics[trim=0 45 0 0, clip, width=\linewidth]{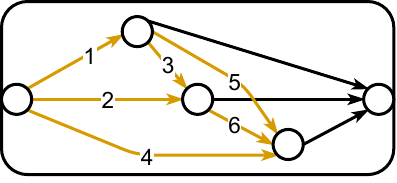}
	\end{minipage}
	\hspace{-0.2cm}
	\caption{
		The NAS-Bench 201 candidate operations and cell design. There are three intermediate nodes and six choices among the available five operations, once for each numbered orange edge.
	}
	\label{fig_p4_nb201}
\end{figure}

In the popular NAS-Bench-201 benchmark \cite{dong2020nasbench201},
architectures are defined by the design of a building block (cell) that is stacked multiple times to create a network.
The cells differ by their chosen candidate operations, which are placed on each of the six marked edges in Figure~\ref{fig_p4_nb201}.
Thus there are $5^6 = 15625$ topologies, with paths of different lengths.
We evaluate the conditional super-network weights on three search space subsets of increasing difficulty:

\begin{enumerate}
	\item {
		All operations are available. Finding above-average models is easy since many networks contain several Zero or Pooling operations and thus perform poorly.
	}
	\item {
		The Zero operation has been removed. The search space thus contains $4^6 = 4096$ architectures. Since most poorly-performing networks are not part of this search space, it is more difficult to find above-average ones.
	}
	\item {
		Only the 1$\times$1 and 3$\times$3 Convolutions remain, reducing the search space to just the $2^6 = 64$ architectures that make up most top-performers in both other search spaces.
		Since all candidates perform well, finding the best architectures in this subset is the most difficult.
	}
\end{enumerate}

Implementing the added super-network weights in this search space is not as straightforward as depicted in Figure~\ref{fig_p4_aw}. Since there are parallel paths in a cell, choosing which other candidates to depend on is ambiguous.
With respect to Figure~\ref{fig_p4_nb201}, we have implemented the meaning of ''prior'' in the following way: The candidates on Paths~1,~2, and~4 have no dependency and are thus never split.
Candidates on Paths~3 and~5 depend on Path~1. Those on Path~6 depend on Paths~2 and~3 and therefore split twice.
The default cells have $6\cdot2=12$ candidates with operation weights (Zero, Skip, and Pooling do not have any and therefore always perform the same function).
Due to the splitting, the total number of weight sets is increased to $3\cdot2\cdot5^0 + 2\cdot2\cdot5^1 + 1\cdot2\cdot5^2 = 76$.
The super-network structure is detailed in the Appendix.

\subsubsection*{NAS-Bench-Macro}

The recent NAS-Bench-Macro benchmark \cite{nbmacro} lists the test accuracy of 6561 fully sequential networks evaluated on CIFAR10.
Their design is inspired by the MobileNet V2 family \cite{sandler2018mobilenetv2}, a popular starting point for modern NAS search space designs. After starting with a 3$\times$3 Convolution layer, one of three available candidates has to be chosen for each of the eight subsequent layers ($3^8 = 6561$).
The available candidate operations are an inverted bottleneck blocks with kernel size 3$\times$3 and expansion ratio 3, another block with kernel size 5$\times$5 and expansion ratio 6, and a skip connection.
The networks have an average accuracy of roughly 90.4\%, with the best network achieving 93.13\%.
Ordinarily, there exist $8\cdot2=16$ weight sets (Skip does not have weights). Splitting increases that to $1\cdot2\cdot3^0+7\cdot2\cdot3^1=44$ (the weights in the first network layer do not depend on any previous layer).

\begin{figure*}[t]
	\begin{subfigure}{0.31\linewidth}
		\includegraphics[trim=10 0 10 40, clip, width=\linewidth]{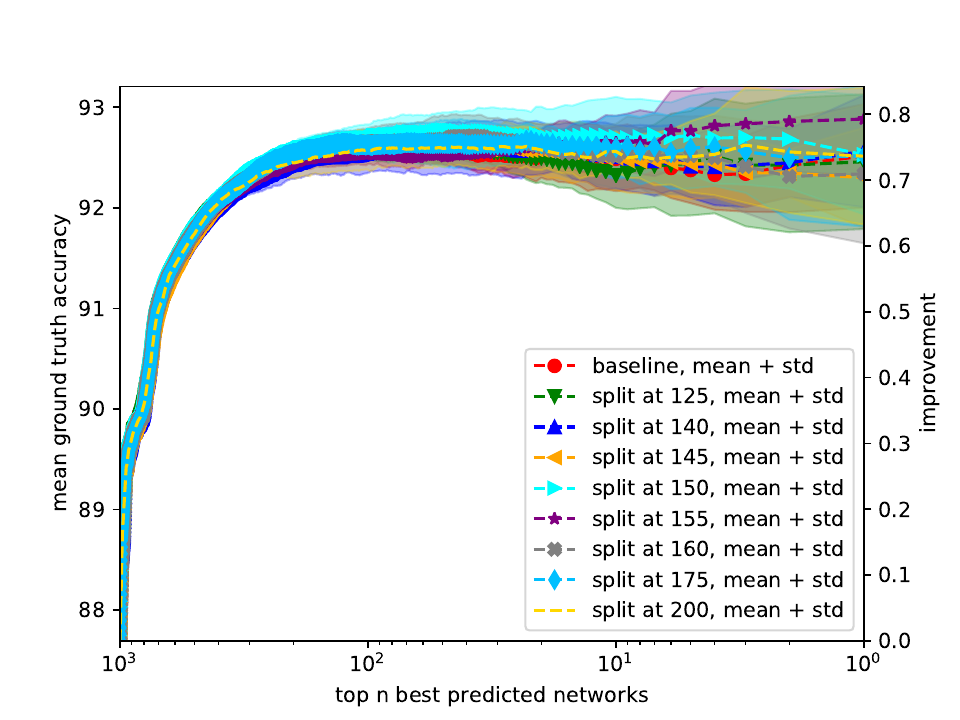}

		\includegraphics[width=\linewidth, trim=2 7 28 40, clip]{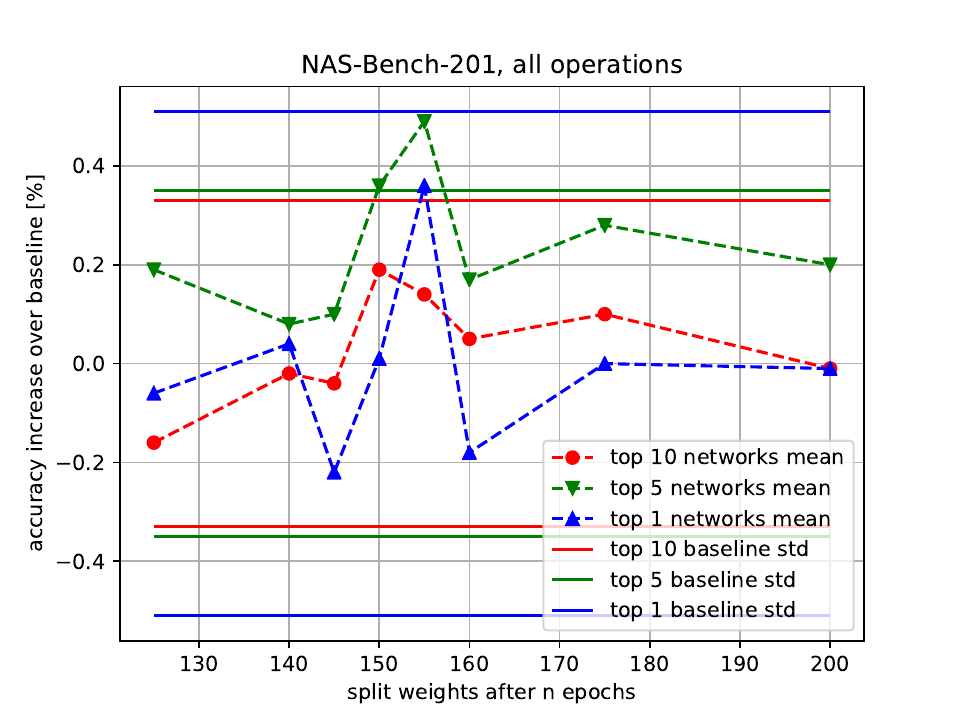}
		\caption{
			All operations
		}
		\label{fig_p4_exp_nb201:m}
	\end{subfigure}
	\hfill
	\begin{subfigure}{0.31\linewidth}
		\includegraphics[trim=10 0 10 40, clip, width=\linewidth]{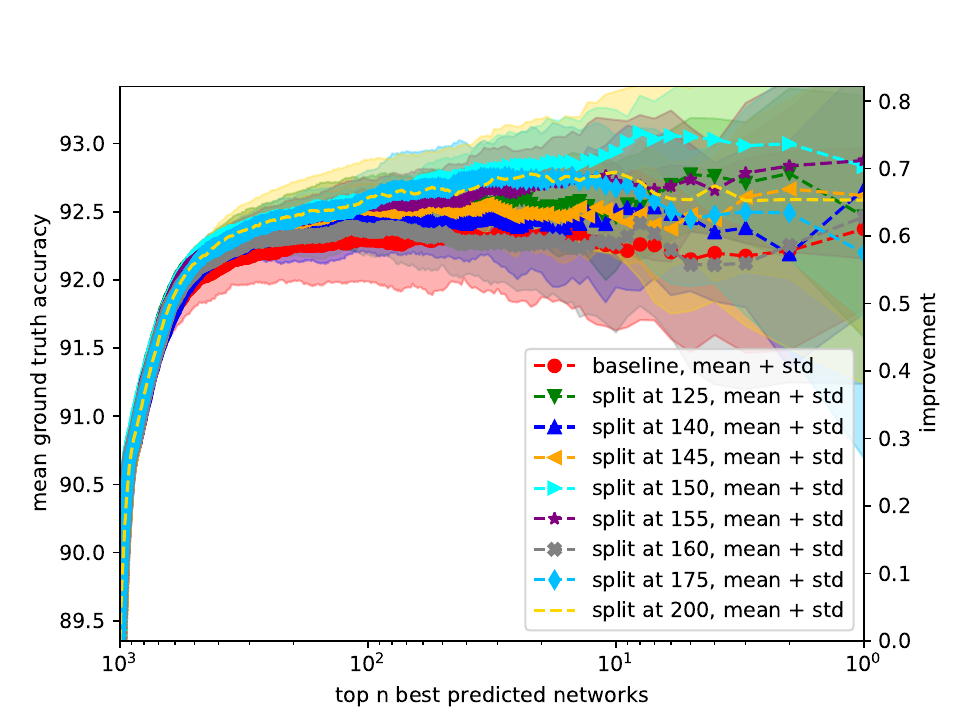}

		\includegraphics[width=\linewidth, trim=2 7 28 40, clip]{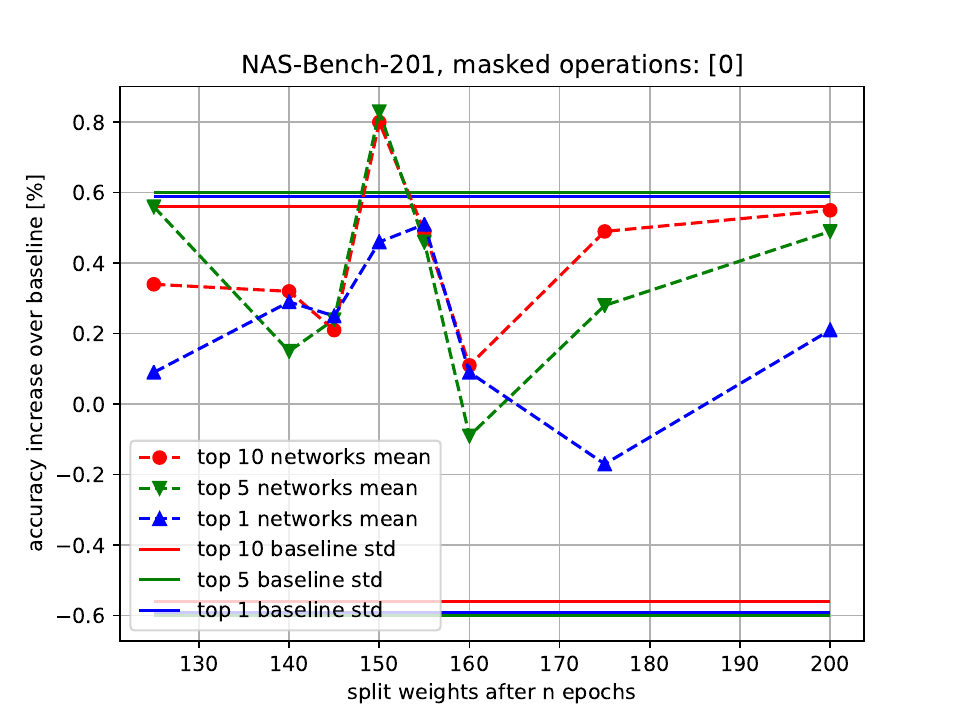}
		\caption{
			No Zero
		}
		\label{fig_p4_exp_nb201:m0}
	\end{subfigure}
	\hfill
	\begin{subfigure}{0.31\linewidth}
		\includegraphics[trim=10 0 10 40, clip, width=\linewidth]{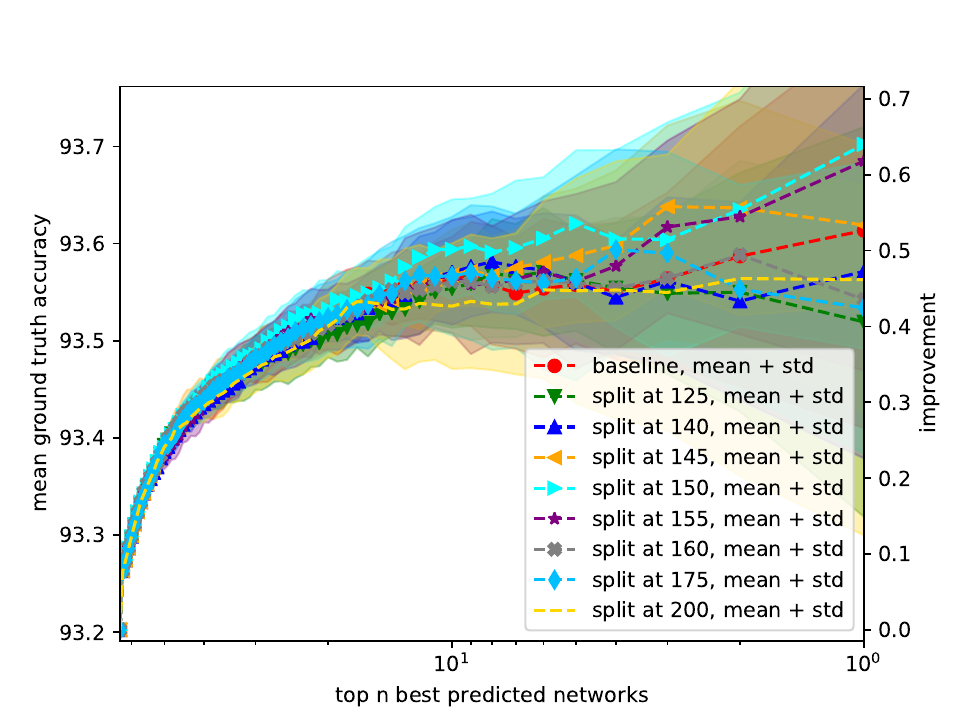}

		\includegraphics[width=\linewidth, trim=2 7 28 40, clip]{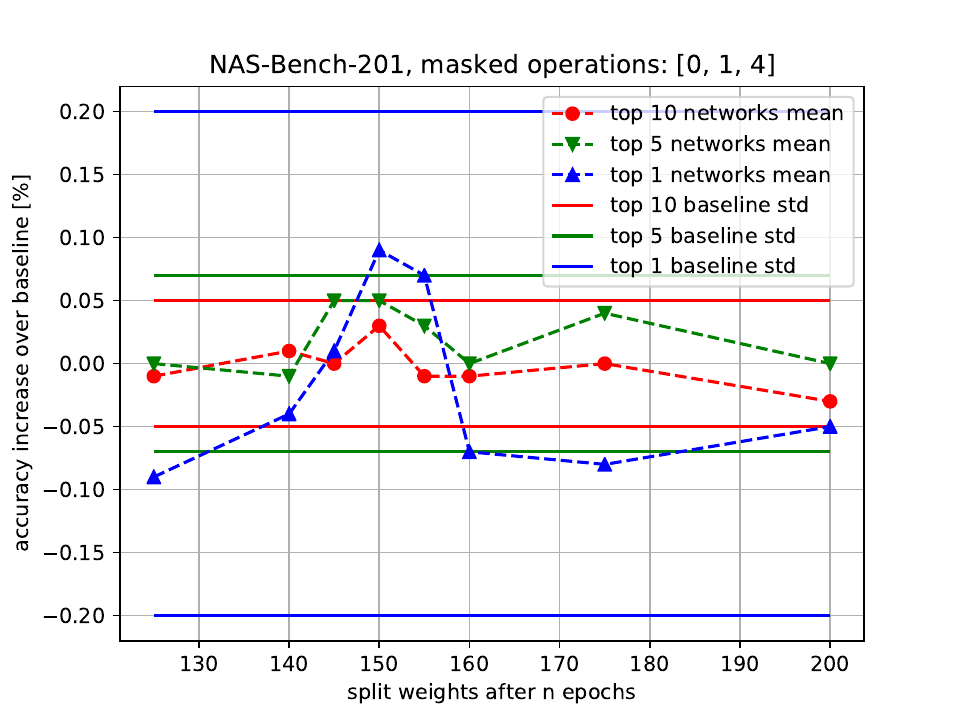}
		\caption{
			Only Convolutions
		}
		\label{fig_p4_exp_nb201:m014}
	\end{subfigure}
	\vskip-0.0cm
	\caption{
		Visualized results of splitting weights in the NAS-Bench-201 subsets, averaged over ten independent trials.
		There is a visible improvement when splitting at around 150 epochs of training.
		\textbf{Top row}: As the super-network predictions are used to remove bad candidate architectures from the search space (x-axis, left to right), the average accuracy of the remaining ones increases (left y-axis).
		The thus increasing improvement value (right y-axis) highlights the benefit of using super-network-based predictions over just picking a random architecture, which has an expected improvement value of~0.
		\textbf{Bottom row}: An overview of how splitting the weights in a specific epoch (x-axis) affects the super-networks top-\{1, 5, 10\} selection, as an absolute improvement over the baseline.
		Fluctuations close to zero are typical artifacts from the randomness of the super-network training.
	}
	\label{fig_p4_exp_nb201}
\end{figure*}

\subsection{Evaluation metrics}
\label{p4_method_eval}

Even though the super-network-based predictions are often used in multi-objective optimization, we simplify the problem by considering only the network accuracy.
The different objectives are generally measured or predicted independently (e.g., accuracy via super-network, latency via {\mtab}), so that improving them in isolation still benefits their combined application.

For a comprehensive evaluation, we first let the trained super-networks rank all architectures in the test set $\mathcal{A}_{test}$.
Since detailed network statistics are known, the average ground-truth accuracy of the top-N selected architectures can be compared.
To better understand the scale of the improvement gained from using a super-network, we also provide a normalized improvement value. It is 0 for the average and 1 for the maximum network accuracy in $A_{test}$.
Both metrics can be seen in the experimental results in Figures~\ref{fig_p4_exp_nb201} and~\ref{fig_p4_exp_nbm}.

\section{Experimental evaluation}
\label{p4_exp}

This section evaluates the effect of adding conditional super-network weights with respect to the regular super-network baseline.
Results are consistently averaged over ten independently trained super-networks, both for their architecture selections in Section~\ref{p4_exp_search} and the resource consumption analysis in Section~\ref{p4_exp_resources}.
Training details are listed in the Appendix.

\subsection{Search results}
\label{p4_exp_search}

\paragraph{NAS-Bench-201}
The results of splitting weights in the multi-path cell-based NAS-Bench-201 super-networks are visualized in Figure~\ref{fig_p4_exp_nb201}.
A fascinating property that all search space subsets have in common is an improvement window when splitting at around 150 epochs of training.
If the timing for weight splitting is just right, the super-networks make notably better suggestions on which networks to select, resulting in improved average accuracy.

As seen in Figure~\ref{fig_p4_exp_nb201:m}, The baseline in the full search space selects networks that achieve around 92.2\% accuracy and therefore improves considerably over the space average of around 87.8\%.
This is also apparent in the improvement value, which is at 0.7. The selected networks are therefore already close to optimal, with little room for improvement.
Nonetheless, splitting weights at around 150 epochs enhances the architecture selection further.
The best value is achieved when splitting the weights after 155 epochs of training, resulting in an average improvement value of around 0.8 for the top-1 selected architectures.

For the super-networks in the No Zero search space, this improvement peak is even more pronounced. As visualized in Figure~\ref{fig_p4_exp_nb201:m0}, the average accuracy changes of the top-5 and top-10 selections reach up to 0.8\%.
The selected networks are considerably better than those of the baseline if the super-network weights are split between 125 and 155 epochs of training.
A notable difference to the full search space is that all super-networks have a wider standard deviation. In the absence of the Zero candidate operation, which is not present in any of the best models, the super-networks are less certain which architectures are relatively better.
Nonetheless, splitting at 145 epochs of training can raise the improvement value from around 0.61 to 0.72, much closer to the optimal network performance. The average accuracy of the top-1 selected networks is improved from 92.3\% to 92.8\%.

Finally, the Only Convolutions search space is the only one where the top-N-selected networks never improve over the baseline by more than one standard deviation. As seen in Figure~\ref{fig_p4_exp_nb201:m014}, the reason is primarily that the baseline's standard deviation is huge.
Another interesting aspect of this search space is that even random sampling results in 93.2\% average network accuracy, much higher than for the other search spaces. Finding the best networks here proves more challenging.
Nonetheless, an increase of the selected top-1 network improvement value from around 0.53 to 0.64 is still obtainable. The average accuracy of the selected networks increases from 93.6\% to 93.7\%.

Interestingly, despite notable improvement peaks such as in Figure~\ref{fig_p4_exp_nb201:m}, the measured Kendall's Tau ranking correlation values change only marginally (see the Appendix).

\begin{figure}
	\vskip -0.0in
	\begin{center}
		\includegraphics[trim=10 0 10 40, clip, width=0.62\linewidth]{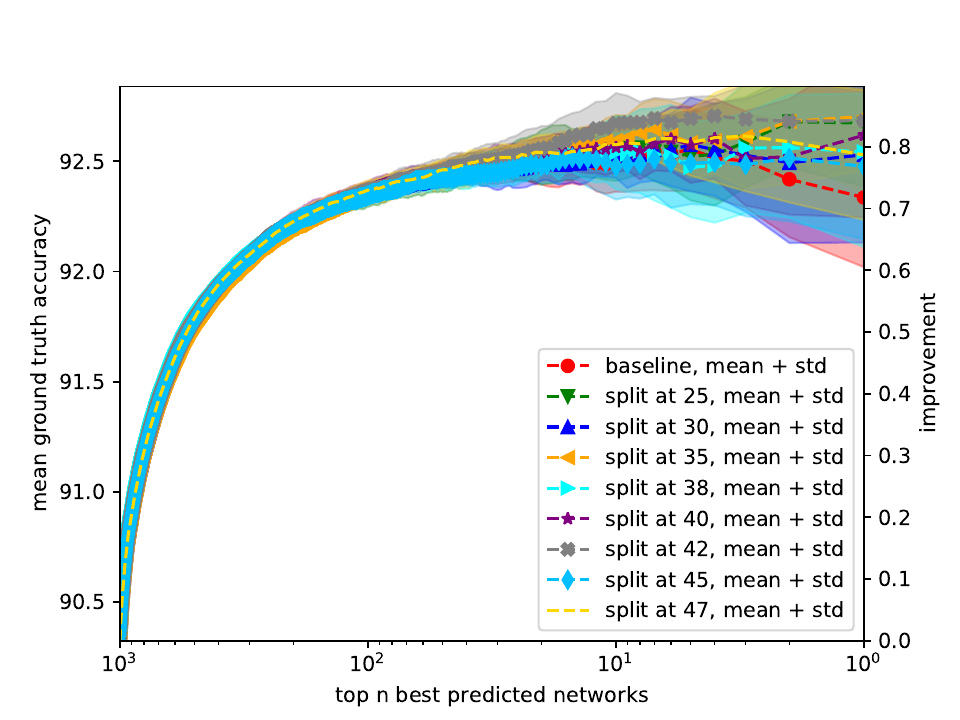}
		\includegraphics[width=0.62\linewidth, trim=2 7 28 40, clip]{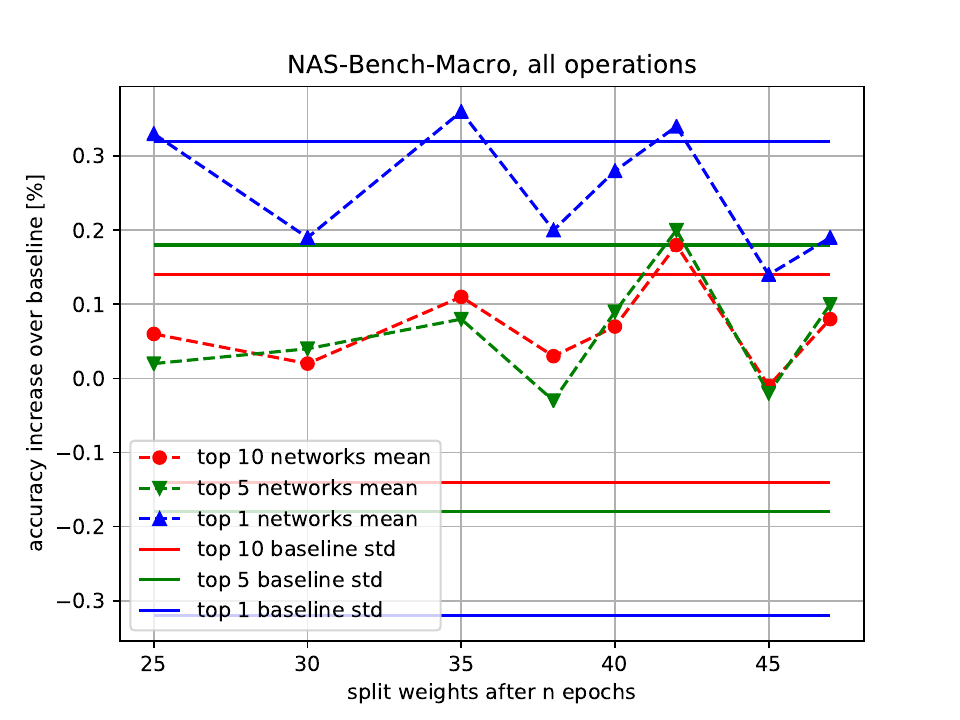}
		\caption{
			Visualized results of splitting weights in NAS-Bench-Macro, averaged over ten independent trials. Since the baseline networks poorly select the top-1 architectures (top image, red), many split variants do better. However, there is little structured effect on the top-10 or top-5 selections.
		}
		\label{fig_p4_exp_nbm}
	\end{center}
\end{figure}

\paragraph{NAS-Bench-Macro}
As seen in Figure~\ref{fig_p4_exp_nbm}, the effect of splitting the network weights is somewhat incomparable to any NAS-Bench-201 search space.
Most notably, the top-1 selected networks are considerably better than those of the baseline at any point. To a lesser degree, the same is true for the top-5 and top-10 selected networks.
The most outstanding epoch for splitting is 42, which is already close to the end of training. At this time, all top-N selections improve by more than one standard deviation over the baseline.

\subsection{Resource analysis}
\label{p4_exp_resources}

\begin{table*}[ht]
	\small
	\hfill
	\begin{minipage}{0.55\linewidth}
		\begin{tabular}{l cc cc cc}
			\toprule
			& \multicolumn{6}{c}{NAS-Bench 201} \\
			\cmidrule(r){2-7}
			& \multicolumn{2}{c}{full} & \multicolumn{2}{c}{no Zero} & \multicolumn{2}{c}{only Conv.}  \\
			\cmidrule(r){2-3}
			\cmidrule(r){4-5}
			\cmidrule(r){6-7}
			&
			\multicolumn{1}{c}{time} & \multicolumn{1}{c}{GPU} &
			\multicolumn{1}{c}{time} & \multicolumn{1}{c}{GPU} &
			\multicolumn{1}{c}{time} & \multicolumn{1}{c}{GPU} \\
			\cmidrule(r){1-1}
			\cmidrule(r){2-3}
			\cmidrule(r){4-5}
			\cmidrule(r){6-7}
			baseline      &     $1862$ &     $1895$ &     $2193$ &     $1895$ &     $2783$ &     $1895$ \\
			\cmidrule(r){1-1}
			\cmidrule(r){2-3}
			\cmidrule(r){4-5}
			\cmidrule(r){6-7}
			split at 125  &  $+17.7\%$ &   $+1.2\%$ &   $+9.6\%$ &   $+0.6\%$ &   $+2.5\%$ &   $+0.0\%$ \\
			split at 140  &  $+14.6\%$ &   $+1.2\%$ &   $+9.9\%$ &   $+0.6\%$ &   $+2.2\%$ &   $+0.0\%$ \\
			split at 145  &  $+17.1\%$ &   $+1.2\%$ &   $+7.9\%$ &   $+0.6\%$ &   $+2.6\%$ &   $+0.0\%$ \\
			split at 150  &  $+13.8\%$ &   $+1.2\%$ &   $+8.1\%$ &   $+0.6\%$ &   $+3.0\%$ &   $+0.0\%$ \\
			split at 155  &  $+14.8\%$ &   $+1.2\%$ &   $+8.9\%$ &   $+0.6\%$ &   $+2.1\%$ &   $+0.0\%$ \\
			split at 160  &  $+13.0\%$ &   $+1.2\%$ &   $+7.5\%$ &   $+0.6\%$ &   $+2.3\%$ &   $+0.0\%$ \\
			split at 175  &  $+11.0\%$ &   $+1.2\%$ &   $+7.1\%$ &   $+0.6\%$ &   $+2.1\%$ &   $+0.0\%$ \\
			split at 200  &   $+9.4\%$ &   $+1.2\%$ &   $+5.7\%$ &   $+0.6\%$ &   $+1.9\%$ &   $+0.0\%$ \\
			\bottomrule
		\end{tabular}
	\end{minipage}
	\hfill
	\begin{minipage}{0.32\linewidth}
		\begin{tabular}{l cc}
			\toprule
			& \multicolumn{2}{c}{NAS-Bench-Macro} \\
			\cmidrule(r){2-3}
			& \multicolumn{2}{c}{full} \\
			\cmidrule(r){2-3}
			& \multicolumn{1}{c}{time} & \multicolumn{1}{c}{GPU} \\
			\cmidrule(r){1-1}
			\cmidrule(r){2-3}
			baseline      &      $729$ &     $3294$ \\
			\cmidrule(r){1-1}
			\cmidrule(r){2-3}
			split at 25   &   $+0.7\%$ &   $+0.7\%$ \\
			split at 30   &   $-0.9\%$ &   $+0.7\%$ \\
			split at 35   &   $+1.9\%$ &   $+0.7\%$ \\
			split at 38   &   $-0.6\%$ &   $+1.0\%$ \\
			split at 40   &   $-2.0\%$ &   $+0.7\%$ \\
			split at 42   &   $+0.4\%$ &   $+0.7\%$ \\
			split at 45   &   $+1.0\%$ &   $+1.0\%$ \\
			split at 47   &   $-3.0\%$ &   $+0.3\%$ \\
			\bottomrule
		\end{tabular}
	\end{minipage}
	\hfill
	\caption{
		Required training resources of the super-networks different scenarios.
		We list the baseline training time in seconds, and the maximum required GPU memory in MB.
		All variations are listed with their respective relative cost increase over the baseline.
		Each network was trained on a single Nvidia 1080 Ti GPU (11GB VRAM), the results are averaged over ten independent runs.
		The super-networks for NAS-Bench-201 and NAS-Bench-Macro have been trained for 250 and 50 epochs, respectively.
		The measured time is not perfectly reliable due to the random selection of candidate operations during training.
	}
	\label{p4_table_exp_res}
\end{table*}

As described in Section~\ref{p4_method_datasets}, splitting increases the total number of super-network weights significantly.
While the purely sequential NAS-Bench-Macro super-networks only have 2.75 times as many candidate operation weight sets as the baseline (16 to 44),
this factor is 6.3 for the multi-path NAS-Bench-201 super-networks (12 to 76).
As detailed in Table~\ref{p4_exp_resources}, the increase in GPU memory is marginal nonetheless, with a peak of only 1.2\%.
The most memory-costly component of training is the saving of intermediate tensors for backpropagation so that storing additional network weights has little effect.

Estimating changes in the training time is less reliable due to the random selection of candidate operations during training. However, a somewhat consistent trend is that NAS-Bench-201 super-networks require slightly more training time when split early.
Unsurprisingly, search spaces that contain more zero-cost operations (Zero, Skip) have a shorter average training time and are thus affected more strongly by the
splitting-overhead than search spaces without.

\subsection{Ablation study}
\label{p4_exp_ablation}

\begin{figure}
	\centering
	\includegraphics[width=0.61\linewidth, trim=2 7 28 40, clip]{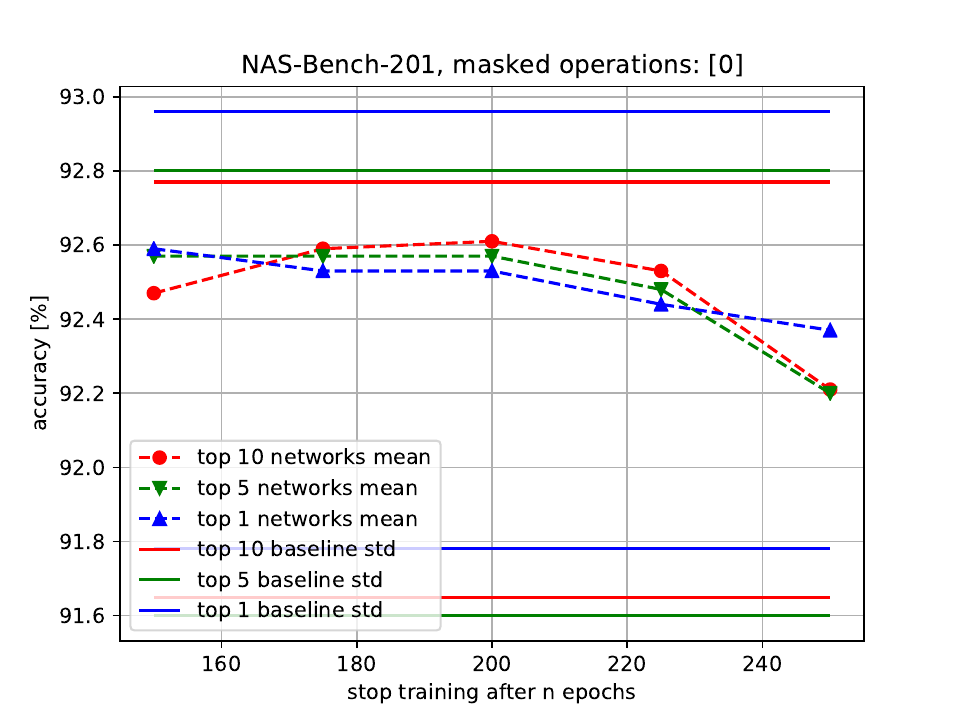}
	\caption{
		Super-networks in the NAS-Bench-201 No Zero search space.
		Simply stopping their training early does not produce the characteristic improvement peak of splitting weights that can be seen in Figure~\ref{fig_p4_exp_nb201:m0}. The results are also well within one standard deviation of the baseline.
	}
	\label{fig_p4_ablation_nb201}
\end{figure}

Since splitting weights comes with a reduced amount of training per weight, it is unclear whether the observed improvement window can be attributed to splitting or the reduced training.
Additional NAS-Bench-201 No Zero super-networks have been trained and evaluated to answer this question. They follow the baseline schedule, except that their training has been suddenly stopped at specific epochs.
As the results in Figure~\ref{fig_p4_ablation_nb201} show, stopping early slightly improved the architecture selection, but not as much or as systematic as the conditional weights.

\section{How to find the improvement window}
\label{p4_window}

As seen in Figure~\ref{fig_p4_exp_nb201}, spitting weights with the correct timing can improve the super-network as a predictor, which results in the selection of significantly better architectures.
However, this evaluation requires prior knowledge of how all selected architectures perform.
If such information was available for real-world problems, there would be no need to apply architecture search.

\begin{figure}
	\vskip -0.0in
	\begin{center}
		\includegraphics[trim=0 0 10 10, clip, width=0.495\linewidth]{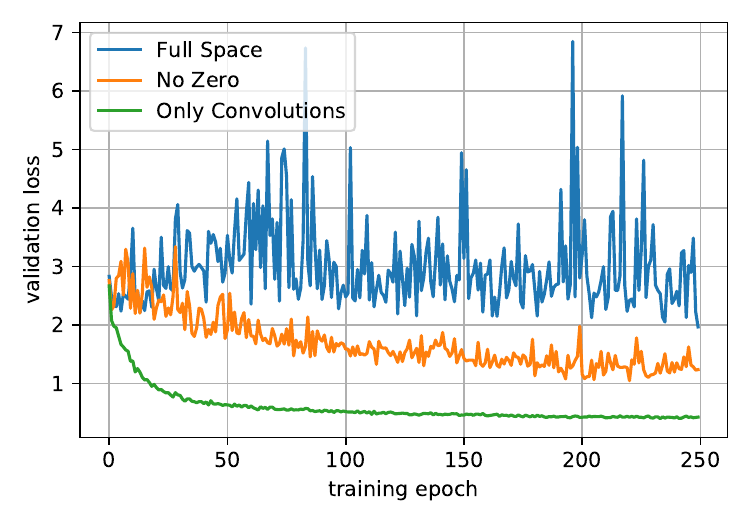}
		\hfill
		\includegraphics[trim=0 0 10 10, clip, width=0.495\linewidth]{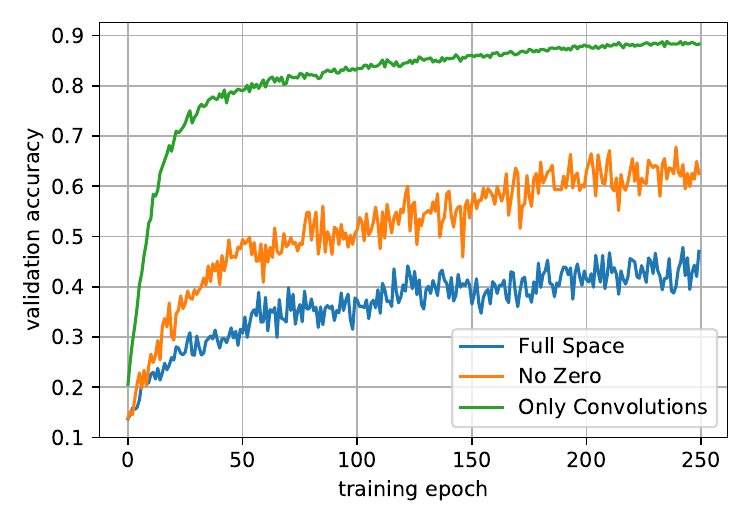}
		\hspace{-0.0cm}
		\vspace{-0.8cm}
		\caption{
			The average validation loss and accuracy during the training of the super-network baselines. While the stochastic selection of candidate operation causes spikes, the averaged network metrics keep improving.
		}
		\label{fig_p4_stats}
	\end{center}
\end{figure}

The true difficulty of the weight splitting method is therefore to predict the improvement window in advance.
Limited to the super-network metrics during training, the clue when to split weights is likely given by validation statistics. However, as visualized in Figure~\ref{fig_p4_stats}, that is not necessarily the case.
The improvement window, situated at around 150 epochs of training, is not matched by any eye-catching property in the validation loss or accuracy. On the contrary, all super-networks keep improving until the end of their training.
This is especially interesting with respect to Figure~\ref{fig_p4_ablation_nb201}, which shows that stopping the training early hardly affects the quality of the subsequently selected architectures.
For now, the connection between the structured improvement windows and any super-network metrics is unclear.

\section{Conclusions}
\label{p4_conc}

In summary, we described the fundamental problem of candidate co-adaptation pressure in super-networks and how conditional weights may reduce its impact.
The presented weight splitting approach enables specializing the weights of every candidate operation to whichever predecessor candidate is currently selected. Even though the total number of network weights is increased multiple times, the training time and memory consumption are hardly affected.

Applying the approach to NAS-Bench-201 results in a curious phenomenon:
splitting weights is most effective when done at around 60\% of the total training epochs and much less so otherwise.
This can be expressed as a window of opportunity when weight splitting should be applied.
With the correct timing, the resulting super-networks are much better in their task of selecting high-accuracy networks.
While the original baseline network selections are somewhat far from the optimal selection possible (the improvement value of 1, they usually achieve between 0.5 and 0.7), the split weights reduce this distance by about a third.
On average, the network accuracy in the Full, No Zero, and Only Convolutions search spaces can be improved by around 0.4\%, 0.8\%, and 0.15\%, respectively.

However, an open question impairs the applicability:
Finding out when to split the weights.
This currently requires prior knowledge about the selected architectures that is only available in NAS benchmarks.
Nonetheless, as the experiments demonstrate, conditional weights are mostly beneficial even when not split optimally.
Future experiments should thus focus on this question and evaluate the approach on a wider range of datasets and training tasks.

{
\bibliographystyle{named}
\bibliography{./bib}
}

\appendix

\section{Network designs}
\label{app_p4_networks}

Tables~\ref{app_p4_table_nb201} and \ref{app_p4_table_nbm} summarize the super-networks for NAS-Bench-201 and NAS-Bench-Macro, respectively.
Both closely follow the design of the original full-sized networks, except for having all candidate operations available.
While NAS-Bench-201 networks usually have five cells of shared topology per stage, our super-network uses only two. This is a typical choice to increase the search efficiency (\cite{zoph2018learning,real2018regularized,pham2018efficient,liu2018darts} and more).

\begin{table*}[ht]
	\begin{tiny}  
		\parbox{.47\linewidth}{
			\centering
			\begin{tabular}{c cc r}
				& & & \\
				& & & \\
				\toprule
				& \multicolumn{2}{c}{input size} &  \\
				\cmidrule(r){2-3}
				cell index & channels & spatial & params \\
				\midrule
				stem & 3 & 32$\times$32 & 464 \\
				0  & 16  & 32$\times$32 & 16,896 \\
				1  & 16  & 32$\times$32 & 16,896 \\
				2  & 16  & 32$\times$32 & 14,464 \\
				3  & 32  & 16$\times$16 & 67,584 \\
				4  & 32  & 16$\times$16 & 67,584 \\
				5  & 32  & 16$\times$16 & 57,600 \\
				6  & 64  & 8$\times$8   & 270,336 \\
				7  & 64  & 8$\times$8   & 270,336 \\
				head & 64 & 8$\times$8  & 778 \\
				\midrule
				sum &  &  & 782,938 \\
				\bottomrule
			\end{tabular}
			\caption{NAS-Bench 201 super-network without additional weights on CIFAR10, using two cells per stage (cells 2 and 5 are fixed reduction cells). As the cell topologies are shared, only six operation choices exist.}
			\label{app_p4_table_nb201}
		}
		\hfill
		\parbox{.47\linewidth}{
			\centering
			\begin{tabular}{c cc r}
				\toprule
				& \multicolumn{2}{c}{input size} &  \\
				\cmidrule(r){2-3}
				layer index & channels & spatial & params \\
				\midrule
				stem & 3  & 32$\times$32 & 928 \\
				0  & 32   & 32$\times$32 & 36,896 \\
				1  & 64   & 16$\times$16 & 87,616 \\
				2  & 64   & 16$\times$16 & 133,184 \\
				3  & 128  & 8$\times$8   & 322,688 \\
				4  & 128  & 8$\times$8   & 322,688 \\
				5  & 128  & 8$\times$8   & 503,936 \\
				6  & 256  & 4$\times$4   & 1,235,200 \\
				7  & 256  & 4$\times$4   & 1,235,200 \\
				head & 256 & 4$\times$4  & 343,050 \\
				\midrule
				sum &  &  & 4,221,386 \\
				\bottomrule
			\end{tabular}
			\caption{NAS-Bench-Macro super-network without additional weights on CIFAR10. In each of the 8 layers, one of the three available operations is chosen for the final architecture.}
			\label{app_p4_table_nbm}
		}
	\end{tiny}
\end{table*}

\section{Training and evaluation details}
\label{app_p4_train}

\begin{table*}[t]
	\begin{small}
	\begin{center}
		\begin{tabular}{l cc}
			\toprule
			& \multicolumn{1}{c}{NAS-Bench 201} & \multicolumn{1}{c}{NAS-Bench-Macro} \\

			\midrule
			Optimizer 				& SGD & SGD  \\
			initial learning rate 	& 0.025 & 0.025 \\
			final learning rate 	& 1e-5 	& 1e-5 \\
			learning rate decay		& cosine & cosine \\
			momentum				& 0.9 & 0.9 \\

			\midrule
			weight decay 			& 3e-4	& 3e-4 \\
			weight decay applies to BatchNorm 	& no & no \\

			\midrule
			epochs					& 250 & 50 \\
			data input shape		& 3$\times$32$\times$32 & 3$\times$32$\times$32 \\
			batch size				& 256 & 256 \\

			\midrule
			training augmentations	& & \\
			pixel shift		& 4 & 4 \\
			random horizontal flipping	& yes & yes \\
			normalization	& yes & yes \\

			\midrule
			evaluation augmentations	& & \\
			normalization	& yes & yes \\

			\bottomrule
		\end{tabular}
	\end{center}
	\vskip-0.3cm
		\caption{
			Training details of super-networks and the evaluation networks for all NAS-Bench 201 and NAS-Bench-Macro networks. Due to using the same method on the same dataset (Single-Path One-Shot on CIFAR10), the configurations are almost identical. If details are not mentioned (such as gradient clipping or dropout), they have not been used.
		}
	\label{app_p4_table_training}
\end{small}
\end{table*}

The baseline super-networks (without additional weights) are outlined in Appendix~\ref{app_p4_networks}.
These networks are trained following exactly the same protocol, except for the optional weight splitting in specific epochs.
The most important details are summarized in Table~\ref{app_p4_table_training}.
The training protocols follow NAS-Bench-201 \cite{dong2020nasbench201} and DARTS \cite{liu2018darts}.
All models are trained on CIFAR10 \cite{cifar10}, where 5000 images are withheld for validation.
We considered training NAS-Bench-Macro super-networks for 100 epochs, but found no advantage in preliminary tests on the baseline.

It is also noteworthy that the BatchNorm statistics of every architecture within the super-network have to be adjusted by performing 20 forward passes (without computing gradients) just prior to evaluating that specific architecture. This is a standard routine of SPOS \cite{guo2020single} which significantly improves the ranking correlation.

All experiments were performed using PyTorch 1.7.0 on Nvidia 1080 Ti GPUs with driver version 440.64, CUDA 10.2, CuDNN 7605, and random seeds \{0,~...,~9\}.

\section{Ranking correlations}
\label{app_p4_ranking}

We present the measured the ranking correlations between super-network predictions and ground-truth accuracy values for the baseline and all variations in Table~\ref{p4_table_exp_kt}.
We consider the entire 1000 test architectures (KT all) and the top-50 ground-truth subset (KT 50).
While we find that splitting weights improves the correlation in most cases, the performance peak when selecting architectures is not obvious.

\begin{table*}[hb]
	\small
	\hfill
	\begin{minipage}{0.55\linewidth}
		\begin{tabular}{l cc cc cc}
			\toprule
			& \multicolumn{6}{c}{NAS-Bench 201} \\
			\cmidrule(r){2-7}
			& \multicolumn{2}{c}{full} & \multicolumn{2}{c}{no Zero} & \multicolumn{2}{c}{only Conv.} \\
			\cmidrule(r){2-3}
			\cmidrule(r){4-5}
			\cmidrule(r){6-7}
			&
			\multicolumn{1}{c}{KT all} & \multicolumn{1}{c}{KT 50} &
			\multicolumn{1}{c}{KT all} & \multicolumn{1}{c}{KT 50} &
			\multicolumn{1}{c}{KT all} & \multicolumn{1}{c}{KT 50} \\
			\cmidrule(r){1-1}
			\cmidrule(r){2-3}
			\cmidrule(r){4-5}
			\cmidrule(r){6-7}
			baseline      &     0.56 &    -0.06 &     0.46 &     0.14 &     0.56 &     0.31 \\
			\cmidrule(r){1-1}
			\cmidrule(r){2-3}
			\cmidrule(r){4-5}
			\cmidrule(r){6-7}
			split at 125  &     0.56 &    -0.08
			&     0.50 &     \textbf{0.19}
			&     0.54 &     0.26
			\\
			split at 140  &     0.55 &    -0.11
			&     0.49 &     0.13
			&     0.54 &     0.29
			\\
			split at 145  &     0.56 &    -0.10
			&     0.50 &     0.12
			&     0.58 &     0.34
			\\
			split at 150  &     \textbf{0.57} &    \textbf{-0.04}
			&     \textbf{0.51} &     0.17
			&     \textbf{0.60} &     \textbf{0.36}
			\\
			split at 155  &     0.56 &    \textbf{-0.04}
			&     \textbf{0.51} &     0.16
			&     0.56 &     0.33
			\\
			split at 160  &     \textbf{0.57} &    -0.07
			&     0.48 &     \textbf{0.19}
			&     0.57 &     0.32
			\\
			split at 175  &     \textbf{0.57} &    -0.09
			&     \textbf{0.51} &     \textbf{0.19}
			&     0.56 &     0.31
			\\
			split at 200  &     0.56 &    \textbf{-0.04}
			&     \textbf{0.51} &     \textbf{0.19}
			&     0.54 &     0.27
			\\
			\bottomrule
		\end{tabular}
	\end{minipage}
	\hfill
	\begin{minipage}{0.32\linewidth}
		\begin{tabular}{l cc}
			\toprule
			& \multicolumn{2}{c}{NAS-Bench-Macro} \\
			\cmidrule(r){2-3}
			& \multicolumn{2}{c}{full} \\
			\cmidrule(r){2-3}
			&
			\multicolumn{1}{c}{KT all} & \multicolumn{1}{c}{KT 50} \\
			\cmidrule(r){1-1}
			\cmidrule(r){2-3}
			baseline      &     0.73 &     0.30 \\
			\cmidrule(r){1-1}
			\cmidrule(r){2-3}
			split at 25   & 0.72 &     0.34
			\\
			split at 30   & 0.73 &     0.31
			\\
			split at 35   & 0.72 &     0.34
			\\
			split at 38   &  0.73 &     \textbf{0.36}
			\\
			split at 40   & 0.73 &     0.31
			\\
			split at 42   & 0.73 &     0.34
			\\
			split at 45   & 0.73 &     0.36
			\\
			split at 47   & \textbf{0.74} &     0.26 \\
			\bottomrule
		\end{tabular}
	\end{minipage}
	\caption{
		Kendall's Tau ranking correlation values between super network predictions and ground truth values, computed on all (KT all) and only the top 50 ground truth networks (KT 50).
		The best column-wise values are marked in bold.
		While splitting weights improves the correlation in most cases, the performance peak when selecting architectures is not obvious.
	}
	\label{p4_table_exp_kt}
\end{table*}

\end{document}